\documentclass{article}

\usepackage[usenames,dvipsnames]{xcolor}

\usepackage{microtype}
\usepackage{graphicx}
\usepackage{subfigure}
\usepackage{booktabs} 

\usepackage[breaklinks]{hyperref}

\usepackage[accepted]{mlsys2020}

\usepackage{enumitem}

\usepackage{etoolbox} 

\newtoggle{draft}
\newtoggle{showlabels}
\newtoggle{norevs}

\togglefalse{draft}

\togglefalse{showlabels}

\togglefalse{norevs}

\iftoggle{showlabels}{
}{}

\iftoggle{draft}{
\usepackage[inline]{showlabels} 

\usepackage[colorinlistoftodos, textwidth=1.5cm, textsize=scriptsize]{todonotes}
\setlength{\marginparwidth}{1.5cm}

}{
\usepackage[colorinlistoftodos, textwidth=1.5cm, textsize=scriptsize, disable]{todonotes}
}
\newcommand{\todoB}[1]{}

\newcommand{\red}[1]{{\color{red}#1}}

\iftoggle{norevs}{

\newcommand{\canca}[1]{}
}{

\newcommand{\canca}[1]{\red{\sout{#1}}}
}

\mlsystitlerunning{Enabling Reproducibility and Meta-learning Through a Lifelong Database of Experiments}

\begin{document}

\twocolumn[
\mlsystitle{Enabling Reproducibility and Meta-learning Through a Lifelong Database of Experiments (LDE)}



\mlsyssetsymbol{equal}{*}

\begin{mlsysauthorlist}
\mlsysauthor{Jason Tsay*}{ibm}
\mlsysauthor{Andrea Bartezzaghi*}{ibmzurich}
\mlsysauthor{Aleke Nolte}{rug}
\mlsysauthor{Cristiano Malossi}{ibmzurich}
\end{mlsysauthorlist}

\mlsysaffiliation{ibm}{IBM Research, Yorktown Heights, NY, USA}
\mlsysaffiliation{ibmzurich}{IBM Research Zurich, R\"uschlikon, Switzerland}
\mlsysaffiliation{rug}{University of Groningen, Groningen, Netherlands}

\mlsyscorrespondingauthor{Jason Tsay}{jason.tsay@ibm.com}

\mlsyskeywords{Experiment Metadata, Experiment Tracking, Meta-learning, Machine Learning, AI Developer Tools, Reproducibility}

\vskip 0.3in

\begin{abstract}
Artificial Intelligence (AI) development is inherently iterative and experimental. Over the course of normal development, especially with the advent of automated AI, hundreds or thousands of experiments are generated and are often lost or never examined again. There is a lost opportunity to document these experiments and learn from them at scale, but the complexity of tracking and reproducing these experiments is often prohibitive to data scientists. We present the Lifelong Database of Experiments (LDE) that automatically extracts and stores linked metadata from experiment artifacts and provides features to reproduce these artifacts and perform meta-learning across them. We store context from multiple stages of the AI development lifecycle including datasets, pipelines, how each is configured, and training runs with information about their runtime environment. The standardized nature of the stored metadata allows for querying and aggregation, especially in terms of ranking artifacts by performance metrics. We exhibit the capabilities of the LDE by reproducing an existing meta-learning study and storing the reproduced metadata in our system. Then, we perform two experiments on this metadata: 1) examining the reproducibility and variability of the performance metrics and 2) implementing a number of meta-learning algorithms on top of the data and examining how variability in experimental results impacts recommendation performance. The experimental results suggest significant variation in performance, especially depending on dataset configurations; this variation carries over when meta-learning is built on top of the results, with performance improving when using aggregated results. This suggests that a system that automatically collects and aggregates results such as the LDE not only assists in implementing meta-learning but may also improve its performance. 

\end{abstract}
]



\printAffiliationsAndNotice{\mlsysEqualContribution} 

\section{Introduction}\label{introduction}

The development of data pipelines and solutions based on Artificial Intelligence (AI) is a heavily experimental and iterative process, necessitating the generation of tens or hundreds of experiments in order to develop a single AI model. This is especially true with the advent of automated machine learning~\cite{feurer2015efficient, feurer2020auto, jin}, promising better performance by automating and scaling up the experimental process to hundreds or thousands of potential models. While the majority of these experiments are often thrown away, there is value in tracking and storing information about these experiments in order to learn from them. At the same time, many AI experiments are often not reproducible~\cite{gundersen2017state}, especially by other data scientists. Reproducing or even just tracking AI experiments -- especially a large number of them -- is complex and time-consuming, requiring a data scientist to document multiple artifacts and phases of the experiment including: the used dataset, how the data was split, the selected model and hyperparameters, the result metrics, and managing where all these artifacts are stored and the interdependencies between them. This large amount of additional work required in tracking data, configurations, and other artifacts often does not directly benefit the data scientists who are developing the AI models, so tracking experiments tends to be at best an afterthought if done at all, resulting in accumulated technical debt~\cite{sculley} for the development process. Therefore, a system that automates the experiment tracking and subsequent reproduction and learning across experiments would both reduce the barriers towards tracking the artifacts and enhance the benefits of having a large set of experiment data.

We propose a Lifelong Database of Experiments (LDE) to automatically extract, manage, and analyze metadata from AI experiments. To lower the barrier of additional work, our LDE implementation includes a Python client to automatically extract and interact with artifacts from all phases of AI experiments from datasets to pipelines to each training or inference run. The metadata extracted from these artifacts contain all the information needed to reproduce a given experiment, such as what dataset was used and how it was split or the hyperparameters used by the model during training. All of the experiment metadata are stored by the LDE server for querying and performing analysis across experiments. The interconnected nature of the artifact metadata allows for complex queries that are well-suited for meta-learning, i.e.~observing machine learning algorithms on a variety of tasks and learning how to improve them. The LDE allows for easily analyzing how particular AI pipelines perform on different types of datasets, and also more specific analyses such as how particular parameter configurations perform on particular datasets or the relationship between parameters and different performance metrics, and so on. The LDE server also makes use of meta-learning algorithms to recommend machine learning pipelines when given the metadata of an unseen dataset.

As a demonstration of the capabilities of the LDE, we perform two experiments using reproduced experiment data from an existing meta-learning study~\cite{fusi2018probabilistic} that spans hundreds of datasets and thousands of machine learning pipelines. The first experiment examines the relationship between reproducibility and variability in the performance results from the reproduced experiment. Given the complex and often non-deterministic nature of machine learning, results from experiments tend not to be absolute. We re-run a number of experiments using the same datasets and pipelines but vary the number of repeated trials and dataset split configurations, in order to examine how the performance metrics vary for this reproduction with respect to the reported ones. We find that, while repeated trials on the same data do not seem to significantly vary performance, dataset split configurations significantly vary the results for most of the datasets in our reproduction set. The second experiment leverages the meta-learning features of the LDE to implement two meta-learning algorithms on top of our reproduced metadata. We examine how variability in experiment results affects the performance of these meta-learning algorithms by building our algorithms on top of two subsets of experimental results: 1) the common case of having a single result per dataset and pipeline combination and 2) having enough data for aggregating multiple results per combination (which is made easier by automated experiment tracking). We find that, as intuition would suggest, both meta-learners perform better with aggregated results. This demonstrates both the importance of collecting as much data as possible and that, when reproducing experiments, it is important to reproduce the context behind them as well, as the performance may unexpectedly vary otherwise. This is especially true when considering meta-learning systems or applications that build on top of these performance results. An automated experiment metadata management system such as the LDE is then an important enabling step towards tracking the metadata necessary towards both accurate reproduction and further meta-learning.

Our main contributions are as follows:
\begin{itemize}[noitemsep]
  \item{
    \textbf{System contributions.} While the concept of AI experiment metadata management systems is not new (e.g.~\cite{Vartak2016, miao2017}), our LDE features the most complete reproducibility-forward architecture design, with automated extraction of interconnected artifact metadata. Additionally, our LDE is the first system that is jointly developed for reproducibility and meta-learning, implementing features that support both activities, while focusing on ease of integration into existing AI workflows. Moreover, the LDE leverages meta-learning algorithms to recommend pipelines given an unseen dataset. The LDE is not strongly tied to any particular technology and we expect its design principles to be applicable towards promoting reproducibility and meta-learning in other experiment or model management systems.
  }
  \item{
    \textbf{Reproducibility and variability contributions.} We use our LDE to reproduce results from a study by Fusi et al.~\cite{fusi2018probabilistic} and examine how various (non-reported) aspects of experimental reproducibility impact result variability. To our knowledge, we are among the first to empirically examine the possible variability when reproducing an AI experiment and the factors that impact this variability. Our results suggest that along with the inherent variability of the reproduced pipelines examined, dataset splitting configurations (despite the same train/test ratio being used as in the original manuscript) have a significant effect on the variability. Based on our results, we suggest best practices to control result variability through data reproducibility.
  }
  \item{
    \textbf{Meta-learning contributions.} By using our LDE, we examine the relationship between experiment result variability and meta-learning performance. To the best of our knowledge, this is the first study that empirically examines this relationship. Our results suggest that meta-learning performance can be generally improved by reducing result variability by aggregating results of experiments. We note that by automatically gathering metadata from experiments and aggregating them along parameter configurations, our system would be able to potentially easily improve any meta-learning algorithm built on top of it.
  }
\end{itemize}

This work is organized as follows: in Section~\ref{relatedwork} we review related work and describe how it influences the design of our system. In Section~\ref{system} we describe our LDE system and its implementation. As example of the capabilities of the LDE, we perform a reproducibility and a meta-learning experiment in Section~\ref{experiment}. Final discussion and conclusions follow in Section~\ref{conclusion}.

\section{Background and Design Principles}\label{relatedwork}

The design of the LDE was greatly influenced by related work in systems that support AI development and previous works on reproducibility and meta-learning in AI.

\subsection{AI Development Tools}


\textit{AI development tools} are systems that aid data scientists in developing AI-based solutions, including training and deploying of AI models. While machine learning (ML) frameworks such as scikit-learn, TensorFlow, or PyTorch are the most direct systems towards supporting data scientists in training AI models, we focus here on tooling that assists in other aspects of the AI development workflow. These include a wide range of tasks, such as validating training data~\cite{polyzotis}, checking for unfairness and biases during training~\cite{fairness360}, and continuous integration of AI models during deployment~\cite{renggli}. Amershi et al.~claim that this complexity and the experimental nature of AI development lead to a need for unified, end-to-end ML systems~\cite{Amershi2019}. ML platforms such as Google's TFX~\cite{baylor_et_al_2017}, Kubeflow, and MLFlow~\cite{zaharia2018accelerating} support reliably producing and deploying ML pipelines into production. The TFX platform integrates a number of components that perform disparate tasks while also focusing on continuous training at production scale. One of these components in TFX is a data validation system described by Polyzotis et al.~\cite{polyzotis} that infers the schema of a dataset and uses it to unit test the data. They note a necessity for co-evolution between the data schema and the machine learning algorithm due to evolving data and domain experts codifying their own knowledge. We note in our work that this sort of co-evolution between artifacts is also vital to fully reproduce a given experiment. In contrast to internal platforms such as TFX that may limit the algorithms or approaches supported, MLFlow~\cite{zaharia2018accelerating} is based on a generic ``open interface'' design that focuses on flexibility. We note that the LDE follows this open approach and adopts similar features, such as having an API and Python client with generic interfaces, together with additional features for known supported types, such as scikit-learn~\cite{scikit-learn} pipelines. In contrast to ML platforms that manage the entire AI lifecycle from end-to-end, the LDE is more similar to model management tools that store, track, and index AI models and related artifacts. One tool of this type is ModelDB~\cite{Vartak2016} which automatically tracks scikit-learn, Spark, and R models by instrumenting code and allows users to view and compare models and pipelines. A similar system with a different scope is ModelHub~\cite{miao2017} which focuses on managing results and snapshots of deep learning models with an optimized parameter storage system. OpenML~\cite{OpenML2013} focuses on building a catalog of datasets and machine learning tasks with the intention of promoting collaboration between data scientists, or ``networked science''. We note that, in contrast with existing management tools, the LDE focuses on reproducibility rather than cataloging or versioning of AI artifacts. This is done through storing metadata around a larger variety of artifacts with more granularity and details than in other systems, while also linking these artifacts together to manage their interdependence.

\subsection{Reproducibility in AI}

Reproducibility of AI experiments in research is an important yet often overlooked aspect of AI development. Gundersen et al.~\cite{gundersen2017state} define reproducibility in AI research as ``the ability of an independent research team to produce the same results using the same AI method based on the documentation made by the original research team''. They further define three degrees of reproducibility: \textit{method}, \textit{data}, and \textit{experiment}, and find in a study of 400 top AI conference papers that most are not reproducible on any degree. We borrow these concepts, especially \textit{data reproducibility}, regarding the documentation of the various datasets used, and \textit{experiment reproducibility}, regarding the approach and environment of the experiment, in the design of our system. Raeder et al.~\cite{raeder} expands on the consequences of data reproducibility by examining how classifier performances vary depending on the cross-validation method used while training. They find that many popular ``canonical'' methods of evaluation, such as a single $k$-fold cross-validation, often differ significantly from ``steady-state'' evaluations, where the cross-validation is run for 500 iterations, and conclude that these popular evaluation methods may not be reproducible. For experiment reproducibility, Pimentel et al.~\cite{pimentel2019} performed a reproducibility study on 1,159,166 Jupyter notebooks and found that only 24\% were runnable and less than 4\% were able to reproduce results. They note that many failures are due to a lack of documentation about package dependencies. As another example of experiment reproducibility, Wang et al.~\cite{wang2020} provide a standardized method to compare the performance of deep learning hardware and software systems. For example, their ParaDnn tool is able to compare the performance of a set of parametrized deep learning models running on Google's TPUv2 vs NVIDIA's V100 GPU. Mattson et al.~\cite{mattson2020} identify reproducibility challenges unique to ML benchmarking, in particular run-to-run variations and the wide variety of software and frameworks available. Their proposed solution is to provide a benchmarking platform called MLPerf that defines standard reference benchmarks and rules for evaluation (as an example of method reproducibility).

\subsection{Meta-learning}

Meta-learning, or ``learning to learn'', studies how ML systems can increase efficiency through experience~\cite{vilalta2002}.
Auto-sklearn~\cite{feurer2015efficient, feurer2020auto} uses meta-learning to warm-start its automated hyperparameter optimization~\cite{feurer2015initializing} by matching meta-features of an unseen dataset to a set of known OpenML datasets. Wistuba et al.~\cite{wistuba2015} seek to improve this usage of meta-learning to initialize hyperparameter optimization for a given dataset by using an adaptive strategy that generates new meta-features to use. Fusi et al.~\cite{fusi2018probabilistic} study the similar meta-learning task of automatically selecting a ML pipeline when given an unseen dataset using probabilistic matrix factorization. This particular collaborative filtering-inspired approach takes advantage of having a large number of existing experiments to better select a pipeline. Our experiments with LDE system, described in Section~\ref{experiment}, partially reproduce the dataset described in this study.

\subsection{Design Principles}

Informed by previous work in AI development tools and the current state of reproducibility and meta-learning, we designed the LDE with the following principles:
\begin{enumerate}[noitemsep]
  \item{
    \textbf{Automated and linked metadata extraction.} While the LDE is not an end-to-end ML platform, we recognize that tracking an AI experiment is complex and touches multiple stages of the AI lifecycle. Given the sheer amount of artifacts generated at various phases, our system aims to extract metadata from these artifacts and to preserve the relationships between them. We also recognize that an important part of our goal is for this extraction to be as transparent and automatic as possible with respect to the end-user, requiring minimal human intervention and changes to existing codebases.
  }
  \item{
    \textbf{Data and Experiment reproducibility.} The LDE aims at reproducing experiments as closely as possible which involves multiple degrees of knowledge about how the experiment was performed. One of our main focuses is to promote data and experiment reproducibility~\cite{gundersen2017state} which requires documenting the datasets used and their configurations and how the experiment itself was implemented along with its hardware and software environment.
  }
  \item{
    \textbf{Meta-learning is complementary.} We observe that gathering a large number of experiments together with detailed information about how they were performed is complementary for training meta-learning algorithms, as a ``free'' benefit. The LDE aims to also support meta-learning algorithms and analyses over stored experimental data by providing meanings to perform complex queries and extract aggregated data.
  }
\end{enumerate}

\begin{figure*}[ht]
\centering
\includegraphics[clip=true, trim=1cm 17.5cm 1cm 0cm, keepaspectratio=true, width=0.8\linewidth]{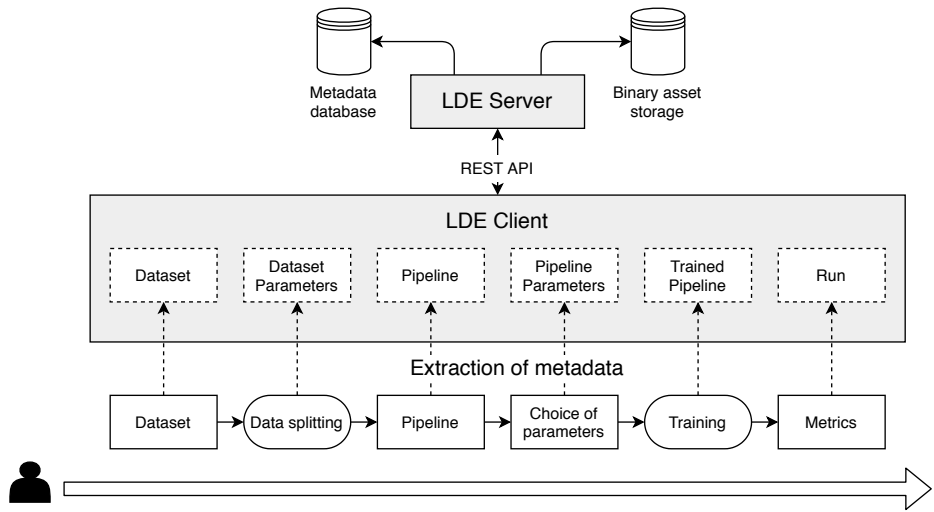}
\vskip -5mm
\caption{Overview of the LDE system: with the help of the LDE Client, metadata is automatically extracted at each step of a typical data scientist workflow and submitted to the LDE Server, which registers it in appropriate databases and provides means for querying and aggregating the collected data.\label{fig:lde_system}}
\vskip -1mm
\end{figure*}

\section{System Overview}\label{system}

We implemented the Lifelong Database of Experiments (LDE) system to assist in automated documentation of AI experiments for reproducibility and meta-learning.
At its core, the LDE consists of two parts: a backend server for storing metadata and binary assets with a REST API interface, and a frontend client to easily issue queries to the backend and automatically extracting and uploading metadata and binaries directly from experiment artifacts. Figure~\ref{fig:lde_system} describes the overall system architecture.
In the following we describe the experiment artifacts that we store metadata for, the database server, and the metadata extraction client.

\subsection{Metadata for AI Experiment Artifacts}\label{sec:metadata}

With the term \textit{AI experiment} we refer to the process of applying a given \textit{method} (i.e.~a Machine Learning / Deep Learning pipeline) to an input \textit{train dataset} in order to solve a specific \textit{task} (e.g.~multi-class classification). Success of this task is measured by one or multiple \textit{metrics} (e.g.~accuracy), calculated against a \textit{test dataset}.

\begin{figure*}[ht]
\centering
\includegraphics[clip=true, trim=0.4cm 0.4cm 0.4cm 0.4cm, keepaspectratio=true, width=0.6\linewidth]{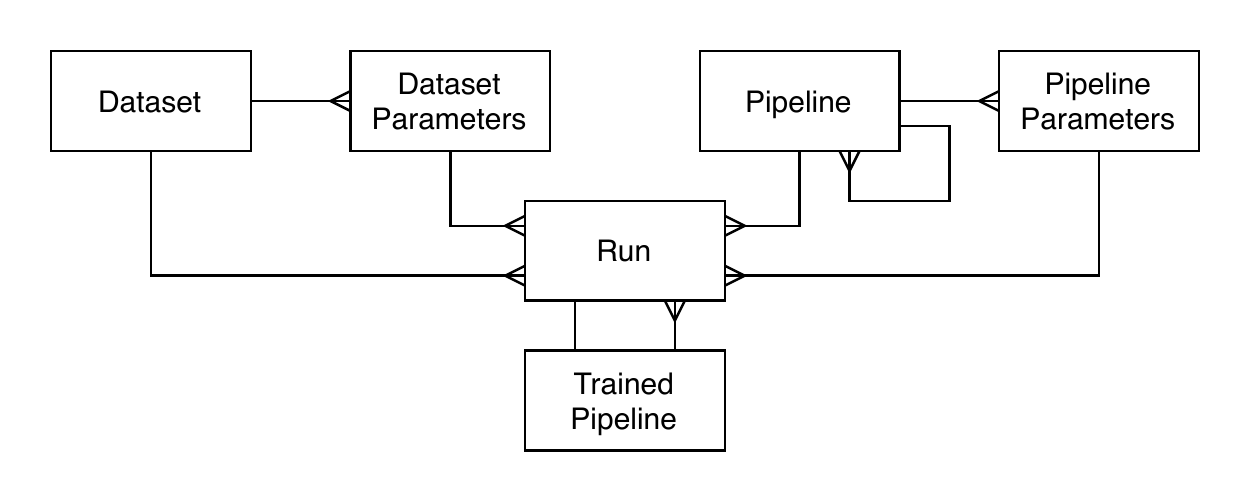}
\vskip -2mm
\caption{Dependencies among metadata artifacts. When links have multiple endings they denote one-to-many relationships.\label{fig:metadata}}
\vskip -2mm
\end{figure*}

Since one of the main goals of the LDE is full reproducibility of AI experiments, every step of the experiment has to be recorded, together with all the details and hyperparameter values determining and influencing its inner operations. We refer to each part composing an experiment as an \textit{artifact}. The artifact \textit{metadata} then refers to the collection of information fully defining an artifact and the context around it. At the time of writing, the types of artifacts that are tracked by the LDE and their relationships are summarized in Figure~\ref{fig:metadata}. Each type of artifact has a corresponding metadata schema and appropriate links to related artifacts. An automatic validation procedure within the LDE server ensures that all artifacts have valid and consistent metadata following a defined schema. Each schema has common metadata including as a uniquely identiying \textit{id} (via standard UUID generation), one of more \textit{authors} to track ownership, and can be assigned multiple \textit{tags}. These common fields aid in organizing entries and can be used when searching in the database, e.g.~to filter the data used for a particular experiment or created by a certain user. The following describes metadata properties that are specific to certain types of artifacts.

\subsubsection{Dataset}\label{sec:dataset}


The \textit{Dataset} artifact describes a single dataset, a collections of data generally used as input for training or testing of models.
The key fields of the \textit{Dataset} metadata are:

{\setlist{nolistsep}
\begin{itemize}[noitemsep]

\item \textit{data schema}: this field contains the schema which contains information about all the dataset features, such as feature type (e.g.~numerical, categorical, text, etc.), valid value ranges (or list of options in case of categorical features), etc. It is in practice an embedded JSON schema definition~\cite{json_schema} which can be used to automatically validate samples of the dataset.

\item \textit{meta features}: used primarily for meta-learning purposes, this field hosts a set of indicators extracted from the actual data.
For example these include the actual number of samples, number of features per sample, number of classes (for datasets meant for multiclass classification tasks), etc., together with statistical properties extracted from each feature. We allow for users to define and add their own meta-features as needed.

\item \textit{target}: optional field describing the default intended task for which the dataset is suitable (e.g.~classification) together with information on features containing target values (e.g.~class annotations).

\item \textit{source}: field describing the origin of the dataset, such as the original location of the dataset on the web (with the additional information required to download the original data, e.g.~a web URL, or an OpenML id).

\end{itemize}}

\subsubsection{Dataset Parameters}\label{sec:dataset_parameters}

The \textit{Dataset Parameters} artifact describes how a given dataset is configured and prepared for a particular training job.
In particular, it contains all necessary information to fully reproduce the results of a given run, including details about how the data is split among train/validation/test datasets (e.g.~splitting method, percentage ratios or full list of indices, random seed, etc.).
There is a one-to-many relationship between \textit{Dataset} and \textit{Dataset Parameters}, i.e.~a certain dataset can potentially be split in multiple ways, each one described by a different \textit{Dataset Parameters} entry.

\subsubsection{Pipeline}\label{sec:pipeline}


\textit{Pipeline} delineates the high-level structure of a given AI model.
It includes a description of all the steps composing the AI pipeline, from input data to final synthesized model. These steps include data preparators, preprocessors, classificators, regressors, etc. Each step should be described in sufficient enough detail for the LDE client to fully reconstruct a pipeline object such that the user can use it to perform more experiments.

Among the others, keys fields of the \textit{Pipeline} metadata are:

{\setlist{nolistsep}
\begin{itemize}[noitemsep]

\item \textit{task type} and \textit{pipeline type}: high-level information about the AI task the pipeline performs (e.g.~classification, regression) and the framework that defines the particular pipeline (e.g.~Scikit-learn, TensorFlow, Lale).

\item \textit{pipeline definition}: step-by-step description of the pipeline, in a format specific to the actual targeted machine learning / deep learning framework.

\item \textit{input data schema}: schema describing the expected input data which can be fed into the pipeline, in a JSON schema format compatible with the \textit{data schema} field of \textit{Dataset} artifacts.

\item \textit{parameter schema}: schema describing the possible hyperparameters exposed by the pipeline, such as their type, range of validity, etc. in JSON schema format.


\end{itemize}

\subsubsection{Pipeline Parameters}\label{sec:pipeline_parameters}


While the \textit{Pipeline} artifact contains an high-level description of the possible hyperparameters, it does not contain actual parameter values, as they can vary between different runs.
Instead, actual parameter values are collected in the \textit{Pipeline Parameters} artifact.
By using together both a \textit{Pipeline} and a \textit{Pipeline Parameters} artifacts it is possible to configure and fully recreate a pipeline. There is a one-to-many relationship between \textit{Pipeline} and \textit{Pipeline Parameters} within the LDE.

\subsubsection{Run}\label{sec:run}


\textit{Run} describes a single computing job, defined as a pipeline applied to input data and resulting in a set of output metrics. Computing jobs include training of a model on input data and inference with a pretrained model on testing data. As such, a \textit{Run} contains links to all previously described artifacts:
{\setlist{nolistsep}
\begin{itemize}[noitemsep]
\item one (or multiple) \textit{Datasets} describing the input data, with related \textit{Dataset Parameters} referring to how the data is split and configured for this particular run;
\item a \textit{Pipeline} describing the architecture of the model, with related \textit{Pipeline Parameters} listing the actual values of hyperparameters used;
\item an optional \textit{Trained Pipeline} (described in Section~\ref{sec:trained_pipeline}), set when performing inference with a previously trained model (obtained with a previously issued and registered run).
\end{itemize}}
The \textit{Run} artifact includes a set of performance metrics (e.g.~accuracy, balanced accuracy, F1 score, etc.) as output of the training or inferencing process, with optionally the inclusion of timing or memory consumption metrics as additional information. For reproducibility purposes, information about the environment, software, and hardware used to perform the computing job are also stored in the metadata.
There is a one-to-many relationship between the combination of input artifacts (\textit{Datasets}, \textit{Dataset Parameters}, \textit{Pipeline}, \textit{Pipeline Parameters}, and \textit{Trained Pipeline}. As a note, the same combination of input artifacts may generate multiple distinct \textit{Run} artifacts due to the potential of obtaining different results due to potential variability or difference in the software/hardware used. In fact, as described in the next section, we demonstrate repeating training jobs to explicitly examine variability.

\subsubsection{Trained Pipeline}\label{sec:trained_pipeline}


For a specific \textit{Run} an optional \textit{Trained Pipeline} entry may be associated as an input or an output.
In general, it describes all resulting post-training artifacts (such as model weights) which an user would like to store (for example to not repeat a long training job) and their locations such that a user is able to download the necessary binary files to instantiate and reuse trained models.
A training \textit{Run} would have a \textit{Trained Pipeline} entry linked as output, while an inference \textit{Run} would have a \textit{Trained Pipeline} artifact referenced as input.
The \textit{Trained Pipeline} metadata contains a link to the original training \textit{Run} such that all input data and settings can be traced back and retrieved.

\subsection{Metadata Storage Server}

The backend of the LDE consists in a standard web service running in cloud connected to a non-SQL database for storing metadata. 
Interaction with the web service is performed through a REST API.
Each artifact schema described in the previous sections has associated API endpoints for standard create-read-update-delete operations.
Metadata schemas are always checked during each query to keep internal data consistency, while still being designed to be versatile and flexible in case there is useful additional data an user would like to store along with the artifacts.
Moreover, the service offers facilities to perform aggregation queries for linked artifacts (for example a pipeline and all parameter configurations) or for meta-learning purposes.
Along with wrapping a database for storing metadata, the LDE server also provides a common object storage for storing and retrieving artifact binary files. A user can use the LDE Python client to upload large local files, such as datasets or trained pipeline binaries, to this object storage, and the location is automatically stored to the corresponding metadata in the LDE database.

\subsection{Artifact Extraction and Reproduction Client}


The frontend of the LDE is a Python client for end-users to automatically extract metadata from AI experiment artifacts and to reproduce existing experiments.
We designed the client to wrap each artifact metadata (as described in Section~\ref{sec:metadata}) in its own Python class.
The instantiation of these classes is a metadata object that can be directly used, modified further, and/or uploaded to the LDE server.
The main goal of the LDE client is to be easy to use and simple to integrate into existing AI development scripts or systems.
Therefore, it provides means to automatically extract metadata when a known type of artifact is provided.
For example, the LDE client understands Pandas dataframes~\cite{pandas} and automatically extracts the dataframe name, data schema (columns and types), and basic meta-features (number of features, instances, etc.) from a given dataframe. The client also directly downloads datasets and metadata from OpenML~\cite{OpenML2013}, feature which was used for performing the experiments described in Section~\ref{experiment}. This metadata includes schema information (``features'' in OpenML) and meta-features (``properties'' in OpenML).
The client also automatically extracts metadata from trained scikit-learn pipelines~\cite{scikit-learn} and Lale pipelines~\cite{lale} including the sequence of operators and hyperparameters used. For unknown types of artifacts, end-users can provide ways for extracting the metadata or manually instantiate the corresponding metadata object and map properties from the given artifact to the metadata object as appropriate. The client is also used to reproduce artifacts that are stored in the LDE, either via recreating the artifact from metadata or directly downloading the stored artifact from object storage or a source link.

\subsection{Reproducibility and Meta-learning Features}

Along with standard storage and querying features for metadata and artifacts, the LDE system has features intended to support reproducibility of AI experiments. In contrast to other experiment managers, we not only store basic information regarding an experiment but additional contextual and lifecycle information in order to support both data and experiment reproducibility~\cite{gundersen2017state}.
To support data reproducibility, we track detailed dataset configuration within \textit{Dataset Parameters} as described in Section~\ref{sec:dataset_parameters}, such as exactly how a given dataset is split for a given experiment, with the exact indices used in splits also stored to exactly reproduce data subsets.
Raeder et al.~\cite{raeder} and our own experiments demonstrate that splitting methods often have a significant impact on reproducing experimental results.
To support experiment reproducibility, as detailed in Section~\ref{sec:metadata} we store the dataset, pipeline, and runtime configuration for a given experiment, with the pipeline setup containing not only the structure of a model but all of the associated hyperparameters chosen, as part of the \textit{Pipeline Parameters} artifact, in order to re-create a pipeline as it was used during an experiment.
The runtime setup such as versions of software frameworks used and information about the hardware is also storable as part of a \textit{Run}.

\textbf{Example.} Having this comprehensive set of metadata stored in the LDE database, by using the LDE client a user is able to fully reproduce an experiment.
Let us assume a user has found a run they are interested in reproducing.
After checking its runtime environment against the information stored in the \textit{Run}, by leveraging the client they can automatically retrieve the \textit{Dataset} metadata and download the actual data in the correct format from the LDE object storage.
Then, the LDE client is able to split the data by following the contents of the \textit{Dataset Parameters} artifact, obtaining the exact splits used in the original run (assuming the dataset is in a supported format).
By using the \textit{Pipeline} metadata together with the hyperparameter values in the \textit{Pipeline Parameters} artifact, the LDE client is able to rebuild the original pipeline object (if it is in a supported format, e.g.~a scikit-learn pipeline) and then train it on the input data.
The obtained performance metrics can finally be compared against the metrics stored in the original \textit{Run}.


As the LDE is intended to store vast quantities of metadata around AI experiments, we also believe that our system is well-suited for meta-learning.
Therefore, we implemented features for analyzing the stored experimental data and building algorithms on top of it. These aggregations are also filterable based on properties in the metadata, such as, in the previous example, only returning pipelines with a specific tag or containing certain parameters.
Our system also itself leverages meta-learning by providing a recommendation endpoint that receives potentially unseen dataset metadata as input and recommends a set of pipelines and pipeline parameters that may perform well.
Examples of built-in functionalities are:
{\setlist{nolistsep}
\begin{itemize}[noitemsep]
\item return all meta-features of a given pool of datasets;
\item given a dataset return the top performing pipelines and pipeline parameters for a given metric;
\item given a dataset return similar datasets;
\item given a new dataset recommend compatible pipelines based on their performance on similar datasets;
\item given a pipeline return the best hyperparameter configurations found in previous runs;
\item given an experiment (as configuration of datasets and pipelines) return all the related runs, for studying variability of the performance metrics.
\end{itemize}}


%
%
%

\section{Experiments}\label{experiment}

To demonstrate the capabilities of the LDE, we perform two empirical experiments using reproduced experimental data:
\setlist{nolistsep}
\begin{enumerate}[nosep]
\item An analysis of reproducibility through the variability of experiment performance, in subsection~\ref{sec:experiment_variability}.
\item An analysis of the relationship between meta-learning performance and result variability, in subsection~\ref{sec:experiment_metalearning}.
\end{enumerate}

\subsection{Methodology}

For our experiments, we decided to reproduce a portion of an existing meta-learning study by Fusi et al.~\cite{fusi2018probabilistic} and store the results of the reproduction in the LDE. The original study collected 553 OpenML~\cite{OpenML2013} datasets, which are intended for binary and multi-class classification problems and contains less than 10,000 samples. The original study used combinations of two pre-processors and fourteen estimators, for a total of 42,000 total possible parameter configurations. Each parameter configuration is run on each dataset, with a maximum runtime limit of 30 seconds, and the normalized accuracy (i.e.~such that perfect accuracy corresponds to 1 and accuracy of random choice corresponds to 0) is reported. The original study reports 79\% of the possible results due to some runs failing for various reasons.

For our reproduced experiments, we run a subset of the original Fusi et al.~experiments and collect 409 OpenML datasets (144 were not available at the time of collection). For each of the 21 combinations of pre-processor and estimator, we run 50 parameter configurations randomly chosen from the original study (a full list of the 21 pipelines is in the appendix). We compute and register various metrics, including normalized accuracy, balanced accuracy, and F1 score. A subset of the original study was used because it was sufficient for our purpose of investigating reproducibility and meta-learning features of the LDE, rather than investing time and resources to fully reproduce the study.

To investigate variability in performance within pipelines and within datasets, we also repeat each experiment nine times. To test dataset variability, a 5-fold stratified cross-validation is run: each dataset is randomly split into five subsets and the model is trained for a total of 5 times using each time one split for testing and the remaining splits for training (resulting in 80\%-20\% train/test split ratio), and the metrics for the 5 different runs are registered in the LDE.
We highlight that only the dataset splitting ratios are documented in the original study~\cite{fusi2018probabilistic}, while other information about the dataset configuration are not, affecting reproducibility of the original results. The splitting method, random seed, and actual indices are stored in the LDE for exact reproducibility. Moreover, to test pipeline performance variability when the data is kept fixed, a given parameter configuration is run five times on the first cross-validation configuration of each given dataset.
In total, following this setup we have collected 1,690,632 successful runs for 187,848 different experiments. Each of the datasets, pipelines, parameter configurations, dataset splits, and runs are stored in the LDE for further analysis. We provide the reproduction set used in our experiments as supplementary material.


\subsection{Performance Variability for Reproduced Experiments}
\label{sec:experiment_variability}

Our first experiment attempts to examine the degree to which our system is able to reproduce results from another study~\cite{fusi2018probabilistic} and to examine the degree of variation for these results. Reproducibility for AI experiments is multi-faceted in that the ability of reproducing an experiment heavily depends on the available reported documentation about the experiment conditions and the environment (software and hardware) in which these have been conducted.
For this experiment, we will focus on \textit{experiment reproducibility} aiming to reproduce the same results when the same AI method is executed on the same set of data~\cite{gundersen2017state}. Given the non-determinstic nature of many AI algorithms and the lack of full documentation, we also wanted to observe the degree to which the reproduced results vary and what factors affect this variation.

We first examine inherent variability by observing and aggregating the performance metric of normalized accuracy across experiments. To examine how normalized accuracy varies for the pipelines executed, we first examine the distribution of accuracy for each of the 21 pipelines in our set. As seen in Figure~\ref{fig:accuracy_bp}, the pipelines are ranked by the average accuracy across all datasets and parameter configurations from most likely to be accurate to least. We compare this to the standard deviation of normalized accuracy within each combination of dataset and parameter configuration for each pipeline. For example, the ``Polynomial $>>$ QDA'' on dataset A using a specific parameter configuration has five different normalized accuracy values (due to different dataset configurations used), so we calculate the standard deviation of these five values and plot them, repeating the process for each combination. As seen in Figure~\ref{fig:std_bp}, the pipelines are ranked by the average standard deviation of accuracy from lowest (least swing) to highest (most swing). In some cases, such as for ``Polynomial $>>$ QDA'' which has both the lowest average and standard deviation for our metric, we expect that pipeline to be consistently low-performing for most datasets. In other cases, we see similar pipelines with similar accuracies on average but very different variability, such as ``Polynomial $>>$ Gradient Boosting'' versus ``Polynomial $>>$ XGradient Boosting'', suggesting that XGradient Boosting estimator perhaps might be chosen over Gradient Boosting not to improve performance but for more consistent performance (which may improve derivative algorithms such as meta-learning).


\begin{figure}[t]
\centering
\includegraphics[clip=true, trim=0cm 0cm 0cm 0cm, keepaspectratio=true, width=\linewidth]{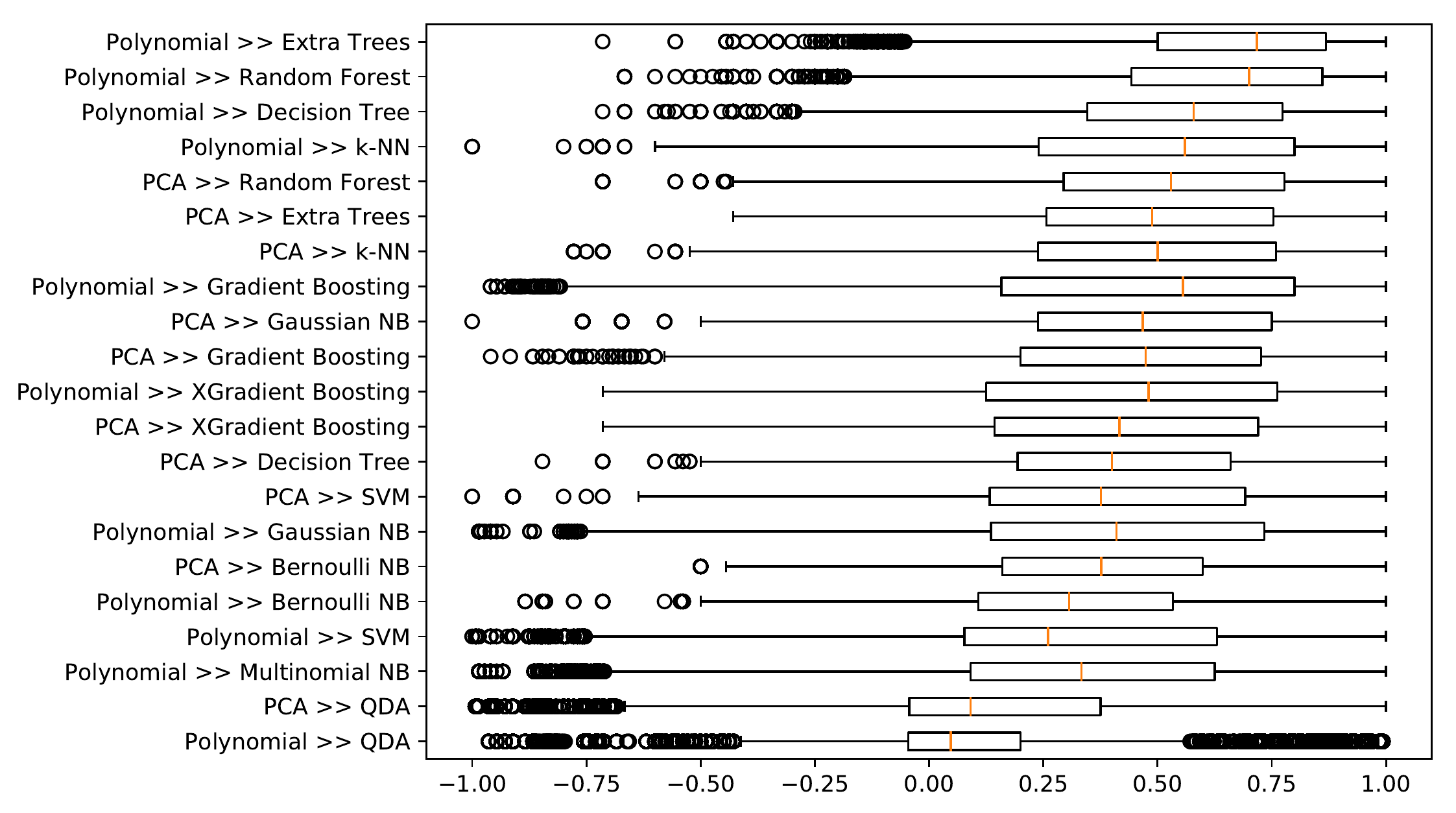}
\vskip -5mm
\caption{Normalized accuracy for each pipeline, sorted by average.\label{fig:accuracy_bp}}
\vskip -2mm
\end{figure}

\begin{figure}[t]
\centering
\includegraphics[clip=true, trim=0cm 0cm 0cm 0cm, keepaspectratio=true, width=\linewidth]{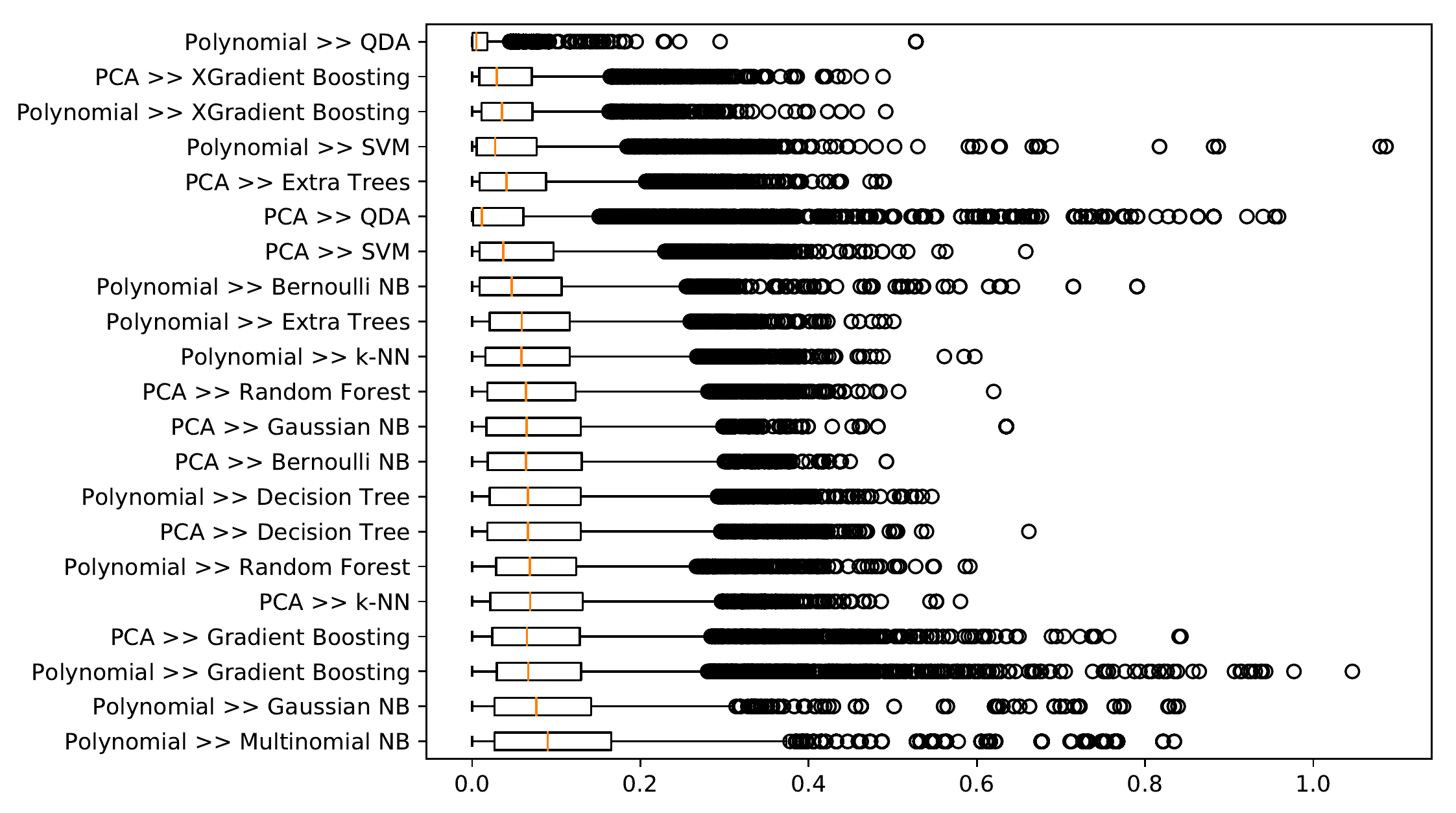}
\vskip -5mm
\caption{Aggregated standard deviation of normalized accuracy for each pipeline, sorted by average.\label{fig:std_bp}}
\vskip -2mm
\end{figure}

While we observe inherent variability, we wish to determine whether there are factors that significantly affect performance variability. For this analysis, rather than comparing individual metrics, we rank the best-performing pipelines and/or parameter configurations per dataset. We use rankings because they are a derivative aggregation from experimental results that are commonly used in meta-learning and are able to test whether rankings differ from each other with statistical significance. As described in the previous section, our system aggregates and returns the top pipelines and top parameter configurations for a given dataset. In the case of top pipelines, all of the results for all parameter configurations for each of the 21 pipelines are averaged. For each dataset in the reproduction set, we examine these rankings when examining two possible reproduction factors: 1) dataset split configurations and 2) repeated trials. Each dataset configuration has the same train/test ratio but different data (though still shuffled and stratified), we generate rankings for each dataset for each dataset configuration for a total of five rankings per dataset. In this way, we are examining whether only varying the dataset configuration will significantly change what pipelines and/or parameter configurations are returned. For the five dataset configuration rankings, we perform the Friedman Test~\cite{Demvsar2006} to test whether the rankings statistically significantly (for a p-value of 0.05) differ. Similarly, we perform five repeated trials on the first configuration to examine if there are other unaccounted factors that may vary results. These five repeated trial rankings are also tested for statistically significant differences using the Friedman Test. The results are summarized in Table~\ref{tbl:rankings}. We see that for most of the 399 datasets, regardless of the number of ranks considered, there is a significant difference in ranking between dataset split configurations but almost no difference between repeated trials. This suggests that the main factor for performance variability for our reproduced set is dataset configuration and there does not otherwise seem to be a missing factor to reliably reproduce performance results. We report normalized accuracy but see similar results when using other metrics: normalized balanced accuracy and F1 score.

Lastly, we perform a similar test against the original study results to see if there is a significant difference in rankings. This is especially important as the original study is a meta-learning study that aimed to automatically select the best-performing pipeline and parameters for an unseen dataset. We would expect that if there is variation in which pipelines and parameters perform the best, then downstream meta-learning would also be affected. We take the original experiment results~\footnote{Acquired from https://github.com/rsheth80/pmf-automl, accessed September 2020} and only consider datasets, pipelines, and parameter configurations that are present in both the original experiment results and our reproduced results for the metric of normalized accuracy (used by the original experiment). We then generate rankings for the top parameter configurations for both the original and reproduced results. Then we use the Wilcoxon Test~\cite{Demvsar2006} to see if the rankings statistically significantly (for a p-value of 0.05) differ and summarize the results in Table~\ref{tbl:rankings}. We see that for most of the 399 datasets, regardless of the number of ranks considered, there is a significant difference in ranking.

\begin{table}[t]\centering
\begin{small}
\begin{tabular}{@{}lrrrr@{}}
Rankings Tested & Top 10 & 20 & 50 & 100 \\
\midrule
Dataset Configs (Pipeline) & 95.5\% & 95.7\% & (na) & (na) \\
Dataset Configs (Param) & 98.5\% & 100\% & 100\% & 100\% \\
Repeated Trial (Pipeline) & 0.5\% & 0.5\% & (na) & (na) \\
Repeated Trial (Param) & 0.3\% & 0.5\% & 1.5\% & 2.0\% \\
\midrule
Original vs Reproduced & 86.0\% & 92.2\% & 91.7\% & 94.0\% \\
\end{tabular}
\end{small}
\vskip -2mm
\caption{Percentage of significantly different rankings (p-value of 0.05). \label{tbl:rankings}}
\vskip -5mm
\end{table}

\subsection{Variability Affecting Meta-learning}
\label{sec:experiment_metalearning}

Our second experiment examines how meta-learning algorithms that are built upon experimental results may be affected by result variability. Intuitively, since many meta-learning algorithms are concerned with identifying best-performing pipelines given an unseen dataset, if variation changes what is considered ``best-performing,'' we also expect the performance of the meta-learning algorithm to be affected. We examine this relationship by implementing meta-learning algorithms on top of the same reproduced result set from the first experiment and compare the performance for two cases: 1) the common case of using a single result per combination of dataset and pipeline configuration and 2) the case where an automated experiment tracker stores multiple results per dataset and pipeline combination. We report meta-learning performance in terms of \textit{regret} or the difference between the best possible normalized accuracy and the normalized accuracy of the pipeline that is recommended by the meta-learning algorithm for the given dataset.

We implement two meta-learning algorithms for our reproduction set: 1) k-Nearest-Dataset (kND)~\cite{feurer2015initializing} which uses dataset meta-features and landmarking to compute similarities between pipeline performances and 2) greedy portfolios~\cite{feurer2020auto} which uses algorithm portfolios rather than meta-features. For this experiment we build each algorithm on a different subset of the reproduced experiment results and observe the impact on regret.

For the first case, we simulate simply having a single result per combination of dataset and pipeline configuration which is common for many meta-learning studies (e.g.~\cite{fusi2018probabilistic, wistuba2015sequential, zhang2008iterative}). In our reproduction set, we simulate this by having our system simply returning the first result to the meta-learning algorithm. We do this instead of selecting a random result to maintain consistency between dataset configurations. In the second case, we allow the meta-learner to have access to the full set of results and return the mean normalized accuracy for all experiments for a given dataset and pipeline configuration. We then compare the regret of both cases for both meta-learning algorithms over 25 iterations for each of the datasets in our reproduced set. The performance of the kND meta-learner is summarized in Figure~\ref{fig:iterative_knn_10_norm_acc} and the greedy portfolios is summarized in Figure~\ref{fig:greedy_portfolios_norm_acc}. Our results suggest that for both meta-learners, the second case of having aggregated results consistently outperforms the common case of a single result. We also repeated this experiment using a different metric of normalized balanced accuracy with similar results.


\begin{figure}[t]
\centering
\includegraphics[clip=true, trim=0cm 0cm 1cm 0cm, keepaspectratio=true, width=\linewidth]{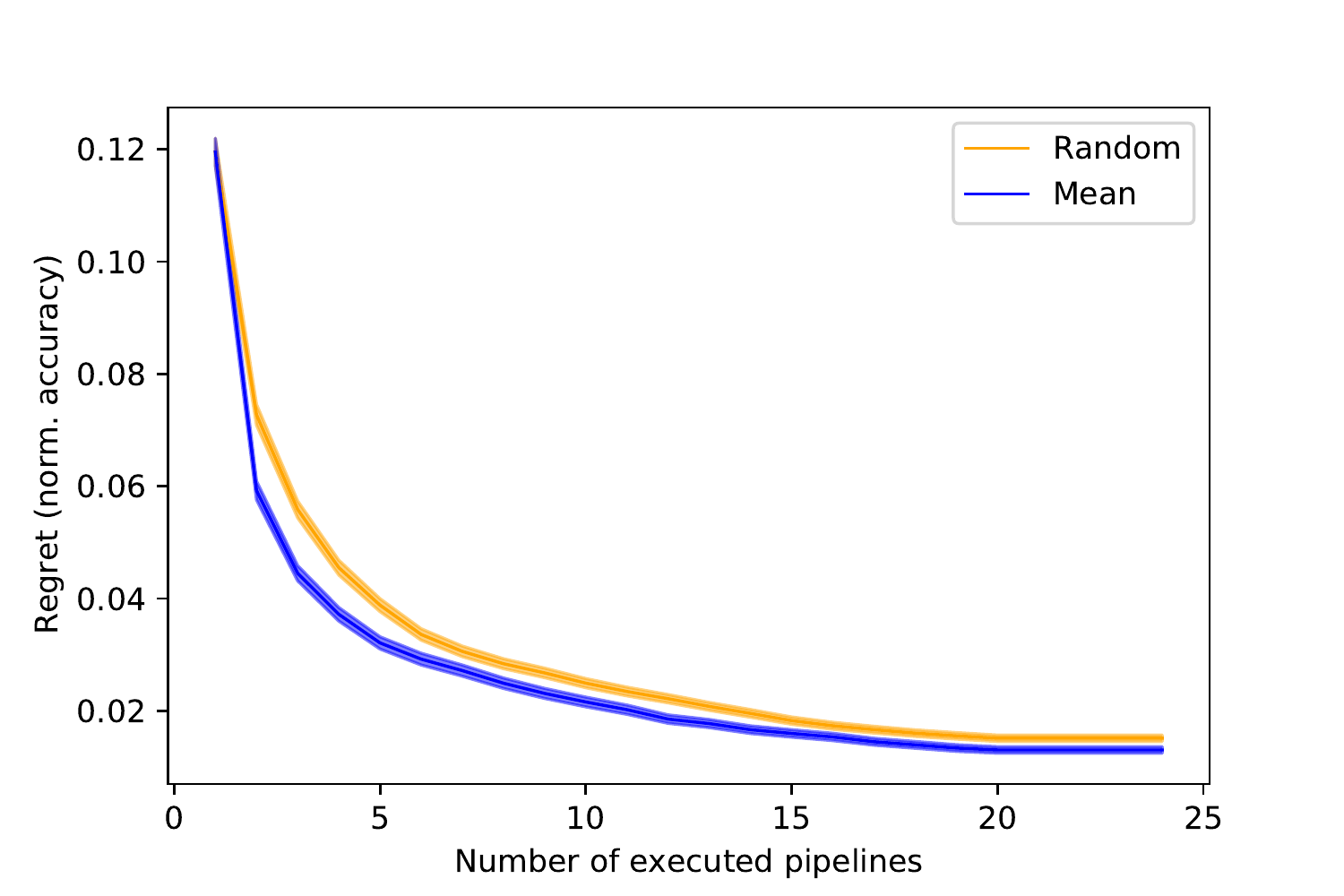}
\vskip -5mm
\caption{Comparative regret of k-Nearest-Dataset (kND) meta-learner with 10 repeats, lower is better.\label{fig:iterative_knn_10_norm_acc}}
\vskip -5mm
\end{figure}

\begin{figure}[t]
\centering
\includegraphics[clip=true, trim=0cm 0cm 1cm 0cm, keepaspectratio=true, width=\linewidth]{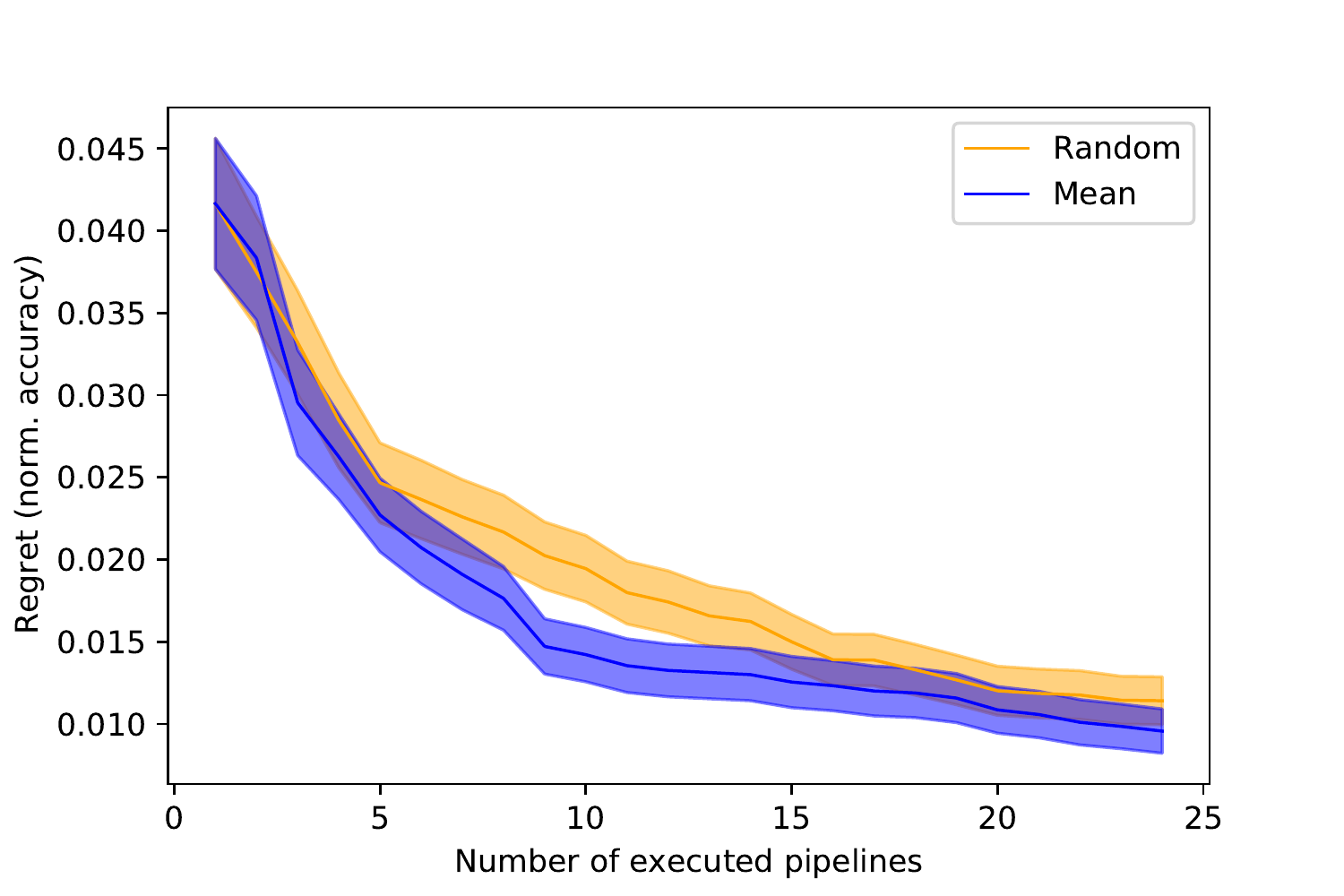}
\vskip -5mm
\caption{Comparative regret of greedy portfolios meta-learner, lower is better.\label{fig:greedy_portfolios_norm_acc}}
\vskip -5mm
\end{figure}

\section{Conclusion}\label{conclusion}

The results of the experiments on reproduced data suggest something many data scientists intuitively understand: that reproducing results of experiments is difficult due to a number of factors contributing to variation in results. However what is less understood is what factors we should be aware of and how they affect systems that build upon experimental results such as meta-learning and active learning systems. Our system allows us to start examining these relationships and we find that along with inherent variation in certain pipelines and datasets, factors that are important in reproducibility such as dataset configuration also play a significant role in result variability. We also find that given this variation, aggregating results seems to give better meta-learning performance. This suggests that systems such as the LDE that automatically collect and aggregate experimental results not only assist in implementing meta-learning systems but may also passively improve them. Although we focus on meta-learning for this paper's experiments, we expect that the metadata collected by our system has similar complementary benefits to other tasks. One possibility is that a training platform could use our system to cache experiments as we store both binary files and enough information to reproduce a partially-run experiment. In summary, we offer the Lifelong Database of Experiments system and its design principles as an example of how to promote reproducibility in AI experiments. The experiments that we perform on reproduced data using our system also demonstrates the potential of using automatically tracking reproducible experimental data for empirical analysis and meta-learning.

\bibliography{main}
\bibliographystyle{mlsys2020}

\appendix
\section{Reproduced Experiments Set}
The reproduced experiment results used in our experiments is available at: https://zenodo.org/record/4077449

\section{Pipelines Used in Experiments}
\begin{table}[h]\centering
\begin{small}
\begin{tabular}{@{}lr@{}}
Pre-processor & Estimator \\
\midrule
PCA & Bernoulli NB \\
PCA & Decision Tree \\
PCA & Extra Trees \\
PCA & Gaussian NB \\
PCA & Gradient Boosting \\
PCA & k-NN  \\
PCA & QDA \\
PCA & Random Forest  \\
PCA & SVM \\
PCA & XGradient Boosting  \\
\midrule
Polynomial & Bernoulli NB  \\
Polynomial & Decision Tree  \\
Polynomial & Extra Trees  \\
Polynomial & Gaussian NB \\
Polynomial & Gradient Boosting  \\
Polynomial & k-NN  \\
Polynomial & Multinomial NB \\
Polynomial & QDA  \\
Polynomial & Random Forest  \\
Polynomial & SVM  \\
Polynomial & XGradient Boosting \\

\end{tabular}
\end{small}
\caption{Pipelines (as a combination of pre-processor and estimator) used in reproduction study (in alphabetical order). \label{tbl:pipelines}}
\end{table}


\end{document}